\documentclass[11pt, a4paper, logo, copyright]{googledeepmind}
\usepackage[authoryear, sort&compress, round]{natbib}
\usepackage[inkscapeformat=png]{svg}

\usepackage[most, breakable, skins]{tcolorbox}

\tcbuselibrary{skins}
\usepackage{lipsum}
\usepackage{tabularx}
\usepackage{afterpage}
\usepackage{booktabs}
\usepackage{subcaption}
\usepackage{enumitem}

\usepackage{makecell}
\usepackage{multirow}
\usepackage{multicol}
\usepackage{array}
\usepackage{epigraph}
\usepackage{float}
\usepackage{listings, listings-rust}
\usepackage{fontawesome5}
\usepackage{amssymb,graphicx}
\usepackage[dvipsnames]{xcolor}
\usepackage{hyperref}
\usepackage[capitalise, noabbrev]{cleveref}
\usepackage{longtable}
\usepackage{graphicx}
\usepackage{ragged2e}
\usepackage{pdflscape}
\usepackage{adjustbox}
\usepackage[section]{placeins}  % float barrier for each section
\usepackage{tikz}
\usetikzlibrary{shapes.geometric, arrows, positioning, fit}
\usepackage{tcolorbox} % For the shaded box around the prompt
\usepackage{listings}
\usepackage{xspace}
\usepackage{caption} % For captions
\usepackage{pifont}

\usepackage{titlesec}
\titlespacing*{\section}{0pt}{*1}{*0.5}
\titlespacing*{\subsection}{0pt}{*0.8}{*0.4}

\newcommand{\ra}[1]{\renewcommand{\arraystretch}{#1}}

\newcommand*{\colorboxed}{}
\def\colorboxed#1#{%
  \colorboxedAux{#1}%
}
\newcommand*{\colorboxedAux}[3]{%
  \begingroup
    \colorlet{cb@saved}{.}%
    \color#1{#2}%
    \boxed{%
      \color{cb@saved}%
      #3%
    }%
  \endgroup
}

\newcommand{\evalname}{\textsl{FACTS}\xspace}

\title{The \evalname Leaderboard: A Comprehensive Benchmark for Large Language Model Factuality}

\correspondingauthor{facts-leaderboard@google.com}

\newcommand{\bGdm}{$\mathbin{\Diamond}$}
\newcommand{\bKaggle}{\ding{168}}
\newcommand{\bResearch}{\ding{171}}
\newcommand{\bCloud}{$\heartsuit$}
\newcommand{\ignore}[1]{}

\author[*,\bResearch]{Aileen Cheng}
\author[*,\bResearch]{Alon Jacovi}
\author[*,\bResearch]{Amir Globerson}
\author[*,\bResearch]{Ben Golan}
\author[*,\bGdm]{Charles Kwong}
\author[*,\bGdm]{Chris Alberti}
\author[*,\bGdm]{Connie Tao}
\author[*,\bResearch]{Eyal Ben-David}
\author[*,\bGdm]{Gaurav Singh Tomar}
\author[*,\bGdm]{Lukas Haas}
\author[*,\bResearch]{Yonatan Bitton}

\author[\bResearch]{Adam Bloniarz}
\author[\bGdm]{Aijun Bai}
\author[\bKaggle]{Andrew Wang}
\author[\bCloud]{Anfal Siddiqui}
\author[\bGdm]{Aravindan Raghuveer}
\author[\bGdm]{Arturo Bajuelos Castillo}
\author[\bResearch]{Aviel Atias}
\author[\bResearch]{Chang Liu}
\author[\bResearch]{Corey Fry}
\author[\bGdm]{Daniel Balle}
\author[\bGdm]{Deepanway Ghosal}
\author[\bResearch]{Doron Kukliansky}
\author[\bResearch]{Dror Marcus}
\author[\bGdm]{Elena Gribovskaya}
\author[\bResearch]{Eran Ofek}
\author[\bGdm]{Honglei Zhuang}
\author[\bResearch]{Itay Laish}
\author[\bGdm]{Jan Ackermann}
\author[\bCloud]{Lily Wang}
\author[\bKaggle]{Meg Risdal}
\author[\bGdm]{Megan Barnes}
\author[\bResearch]{Michael Fink}
\author[\bKaggle]{Mohamed Amin}
\author[\bResearch]{Moran Ambar}
\author[\bResearch]{Natan Potikha}
\author[\bGdm]{Nikita Gupta}
\author[\bResearch]{Nitzan Katz}
\author[\bGdm]{Noam Velan}
\author[\bResearch]{Ofir Roval}
\author[\bResearch]{Ori Ram}
\author[\bGdm]{Polina Zablotskaia}
\author[\bKaggle]{Prathamesh Bang}
\author[\bGdm]{Priyanka Agrawal}
\author[\bGdm]{Rakesh Ghiya}
\author[\bGdm]{Sanjay Ganapathy}
\author[\bGdm]{Simon Baumgartner}
\author[\bResearch]{Sofia Erell}
\author[\bGdm]{Sushant Prakash}
\author[\bGdm]{Thibault Sellam}
\author[\bGdm]{Vikram Rao}
\author[\bGdm]{Xuanhui Wang}
\author[\bGdm]{Yaroslav Akulov}
\author[\bKaggle]{Yulong Yang}
\author[\bGdm]{Zhen Yang}
\author[\bCloud]{Zhixin Lai}
\author[\bResearch]{Zhongru Wu}
\author[\bGdm]{Anca Dragan}
\author[\bResearch]{Avinatan Hassidim}
\author[\bGdm]{Fernando Pereira}
\author[\bGdm]{Slav Petrov}
\author[\bGdm]{Srinivasan Venkatachary}
\author[\bGdm]{Tulsee Doshi}
\author[\bResearch]{Yossi Matias}
\author[\bResearch]{Sasha Goldshtein}
\author[\bGdm]{Dipanjan Das}

\affil[*]{Equal Contribution}
\affil[\bGdm]{Google DeepMind}
\affil[\bResearch]{Google Research}
\affil[\bCloud]{Google Cloud}
\affil[\bKaggle]{Kaggle}

\renewcommand{\comment}[1]{}

\begin{abstract}

We introduce The FACTS Leaderboard, an online leaderboard suite and associated set of benchmarks that comprehensively evaluates the ability of language models to generate factually accurate text across diverse scenarios. The suite provides a holistic measure of factuality by aggregating the performance of models on four distinct sub-leaderboards: (1) FACTS Multimodal, which measures the factuality of responses to image-based questions; (2) FACTS Parametric, which assesses models’ world knowledge by answering closed-book factoid questions from internal parameters; (3) FACTS Search, which evaluates factuality in information-seeking scenarios, where the model must use a search API; and (4) FACTS Grounding (v2), which evaluates whether long-form responses are grounded in provided documents, featuring significantly improved judge models. Each sub-leaderboard employs automated judge models to score model responses, and the final suite score is an average of the four components, designed to provide a robust and balanced assessment of a model’s overall factuality. The FACTS Leaderboard Suite will be actively maintained, containing both public and private splits to allow for external participation while guarding its integrity. It can be found at \url{https://www.kaggle.com/benchmarks/google/facts}.

\end{abstract}

\begin{document}

\maketitle

\section{Introduction}
\label{sec:intro}

Large Language Models (LLMs) have improved dramatically in recent years, yet they continue to generate factually inaccurate information. Indeed, factuality remains one of the most critical and challenging aspects of LLMs. Research in this area can be broadly divided into two distinct scenarios: (1)~factuality with respect to a given context, where a model’s response must be fully grounded in the provided information~\citep[e.g., a document or an image;][]{honovich-etal-2022-true,facts_grounding,tang-etal-2024-minicheck,rashkin2023measuring}, and (2)~factuality with respect to general world knowledge, where a model must accurately answer factoid queries using its internal parameters~\citep{kwiatkowski2019natural,lin2022truthfulqa,chen2023felm} or by using external sources like the web \citep{wei2024measuring,yang2024crag,wei2025browsecomp,vu2023freshllms,mialon2023gaiabenchmarkgeneralai}. Practical use cases would typically rely on both these capabilities (e.g., analyzing financial reports).

While previous works, including our own FACTS Grounding benchmark~\citep{facts_grounding}, each focused on specific capabilities, a comprehensive understanding of an LLM’s factuality requires evaluating its performance across a wider spectrum of tasks. Models that excel at summarizing a provided document may struggle to answer factoid questions from memory, and vice-versa. A robust factuality benchmark should therefore measure a model’s capabilities in multiple contexts, including its handling of different modalities (text, images), knowledge sources (provided context, internal parameters, external search), and response formats.

The FACTS Leaderboard introduced here is designed to address this need by providing a holistic evaluation suite. It aggregates performance across four specialized sub-leaderboards, each targeting a distinct dimension of factuality.
\begin{itemize}
    \item \textbf{FACTS Multimodal} tests a model's ability to combine visual grounding with world knowledge to answer questions about an image.
    \item \textbf{FACTS Parametric} measures the model's ability to use its internal knowledge accurately in factoid question use-cases.
    \item \textbf{FACTS Search} evaluates the practical and increasingly common use case of generating factual responses by interacting with a search tool.
    \item \textbf{FACTS Grounding v2} is an updated version of FACTS Grounding, which tests grounding to a given document, with improved judges.
\end{itemize}

By combining these diverse evaluations into a single suite, we aim to offer a more comprehensive and robust measure of a model's factual reliability, rather than focusing on a narrow set of tasks. 

In what follows, we introduce the four pillars of the FACTS Leaderboard Suite, detail the methodology for each, and explain the aggregation process that yields a single, comprehensive factuality score. We believe this suite offers a nuanced and thorough tool for tracking progress in the ongoing challenge of building factually reliable LLMs.

\begin{table}[t]\centering
\caption{Main results on the FACTS benchmark suite. Values represent Accuracy with 95\% confidence intervals, averaged over both the public and the private datasets. The \textit{FACTS Score} is an aggregate metric representing the average performance across all the four subsets.}\label{tab:main-results}
\begin{tabular}{rlrrrrr}\toprule
\# &Model &FACTS Score &Grounding &Multimodal &Parametric &Search \\\midrule
1  &Gemini 3 Pro    &$68.8$ &$69.0_{\pm{2.1}}$ &$46.1_{\pm{3.6}}$ & $76.4_{\pm{1.8}}$ &$83.8_{\pm{1.6}}$ \\
2  &Gemini 2.5 Pro  &$62.1$ &$74.2_{\pm{2.0}}$ &$46.9_{\pm{3.6}}$ & $63.2_{\pm{2.0}}$ &$63.9_{\pm{2.0}}$ \\
3  &GPT 5           &$61.8$ &$69.6_{\pm{2.1}}$ &$44.1_{\pm{3.5}}$ & $55.8_{\pm{2.1}}$ &$77.7_{\pm{1.6}}$ \\
4  &Grok 4          &$53.6$ &$54.7_{\pm{2.3}}$ &$25.7_{\pm{3.1}}$ & $58.6_{\pm{2.1}}$ &$75.3_{\pm{1.8}}$ \\
5  &GPT o3          &$52.0$ &$36.2_{\pm{2.2}}$ &$39.9_{\pm{3.5}}$ & $57.1_{\pm{2.1}}$ &$74.8_{\pm{2.0}}$ \\
6  &Claude 4.5 Opus &$51.3$ &$62.1_{\pm{3.3}}$ &$39.2_{\pm{3.5}}$ & $30.6_{\pm{2.0}}$ &$73.2_{\pm{1.7}}$ \\
7  &GPT 4.1         &$50.5$ &$45.6_{\pm{2.3}}$ &$40.1_{\pm{3.5}}$ & $51.5_{\pm{2.1}}$ &$64.6_{\pm{2.2}}$ \\
8  &Gemini 2.5 Flash&$50.4$ &$69.9_{\pm{2.1}}$ &$41.0_{\pm{3.5}}$ & $30.7_{\pm{2.0}}$ &$60.0_{\pm{2.2}}$ \\
9  &GPT 5.1         &$49.4$ &$50.0_{\pm{2.3}}$ &$41.8_{\pm{3.5}}$ & $43.2_{\pm{2.1}}$ &$62.4_{\pm{2.1}}$ \\
10 &Claude 4.5 Sonnet \textit{Thinking} &$49.1$ &$61.8_{\pm{2.3}}$ &$31.1_{\pm{3.3}}$ & $29.0_{\pm{1.9}}$ &$74.5_{\pm{1.8}}$ \\
11 &Claude 4.1 Opus &$46.5$ &$54.8_{\pm{2.4}}$ &$33.1_{\pm{3.4}}$ & $33.2_{\pm{2.0}}$ &$65.0_{\pm{1.9}}$ \\
12 &GPT 5 mini      &$45.9$ &$58.3_{\pm{2.3}}$ &$41.5_{\pm{3.5}}$ & $16.0_{\pm{1.5}}$ &$67.9_{\pm{1.9}}$ \\
13 &Claude 4 Sonnet &$42.8$ &$56.1_{\pm{2.3}}$ &$28.6_{\pm{3.2}}$ & $20.4_{\pm{1.8}}$ &$66.3_{\pm{1.9}}$ \\
14 &GPT o4 mini     &$37.6$ &$29.3_{\pm{3.0}}$ &$34.5_{\pm{3.4}}$ & $20.5_{\pm{1.7}}$ &$66.2_{\pm{2.0}}$ \\
15 &Grok 4 Fast     &$36.0$ &$43.1_{\pm{2.3}}$ &$17.7_{\pm{2.7}}$ & $15.8_{\pm{1.6}}$ &$67.3_{\pm{1.9}}$ \\
\bottomrule
\end{tabular}
\label{eval_results_all_slices}
\end{table}

\section{The FACTS Leaderboard}
\label{sec:suite}

The FACTS suite provides a systematic protocol for evaluating LLMs on diverse aspects of factuality. We maintain a live leaderboard tracking performance on the four FACTS benchmarks: Multimodal, Parametric, Search, and Grounding. The leaderboard will remain open to new model submissions.

To mitigate overfitting, only a subset of the prompts will be released publicly, and the remaining prompts will remain private. All model evaluation will be conducted by Kaggle.%\amirg{do we want to address Google's access to these?}  

Detailed metrics for each of the four tasks will be published on dedicated task-specific leaderboards. Since each task measures a different aspect of factuality, analyzing them separately yields the most complete insight. However, to facilitate comparison across benchmarks, the main leaderboard will feature a single holistic performance metric: the average accuracy across all four tasks, where accuracy per task is the average over the public and private sets. We refer to this metric as the ``FACTS Score''.

Table~\ref{tab:main-results} presents the evaluation of proprietary API-based models across the full FACTS benchmark suite (i.e., both private and public sets). For this overview, we report the FACTS Score, as well as the accuracy for each subset, with 95\% confidence intervals.

In the following sections, we describe the four benchmarks, outlining the dataset construction, defining the tailored metrics, and analyzing benchmark-specific outcomes.

\section{FACTS Multimodal}
\label{sec:mm}
\vspace{-4pt}

\begin{table*}[t!]
\centering
\caption{Examples from the FACTS Multimodal benchmark, illustrating the rubric-based evaluation. The autorater uses the rubric to generate two distinct verdicts for factuality and completeness.}
\vspace{-2px}
\label{tab:mm_examples_final}
\scriptsize
\ra{1.3} % Adds vertical spacing to rows
\resizebox{\linewidth}{!}{%
\begin{tabular}{p{3.1cm} p{4.2cm} p{4.5cm} p{2.1cm} p{4.5cm}}
\toprule
\textbf{User Request} & \textbf{Human-Collected Rubrics} & \textbf{Candidate Model Response} & \textbf{Coverage \newline Verdict} & \textbf{ Contradiction Verdict} \\
\midrule

What is this and what year was it introduced? \newline
\includegraphics[width=\linewidth]{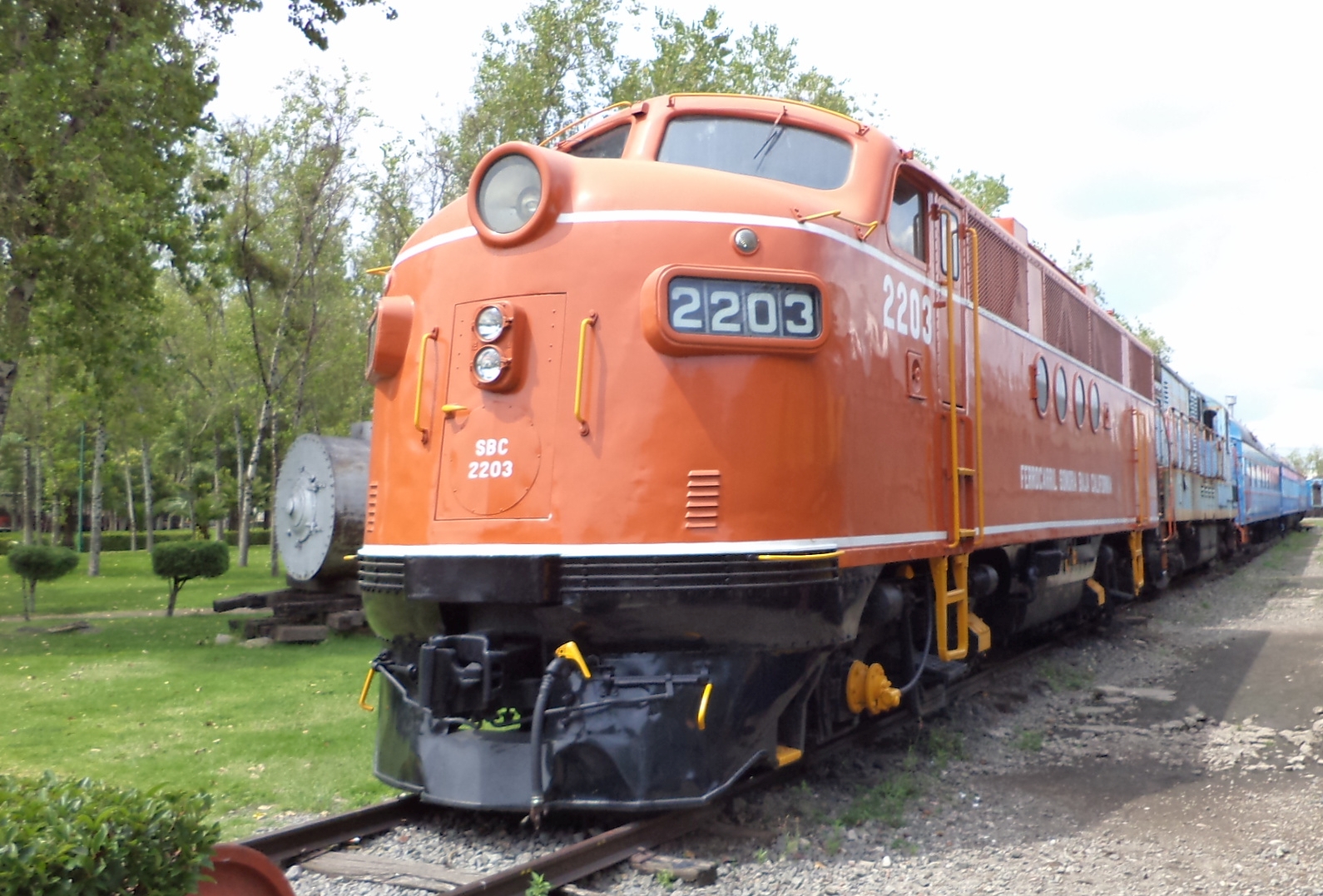}
\tiny{Image: \href{https://commons.wikimedia.org/wiki/File:Locomotora_FSBC_2203.jpg}{Locomotora FSBC 2203} by Es778, \href{https://creativecommons.org/licenses/by-sa/4.0/deed.en}{CC BY-SA 4.0}}
&
\begin{enumerate}[label=\arabic*., leftmargin=*, topsep=0pt, partopsep=0pt, itemsep=0pt, parsep=0pt]
    \item[] \textit{\underline{Essential Facts}}
    \item This is an image of the FC Sonora-Baja California (SBC) 2203 locomotive train.
    \item The FC Sonora-Baja California (SBC) 2203 was introduced in 1949.
    \item[] \textit{\underline{Non-essential Facts}}
    \item The image shows a train at the National Museum of Mexican Railways.
    \item The train in the image has the numbers "2203" across its front.
    \item The image shows an orange train on a sunny day.
\end{enumerate} &
Based on the visual details, this is a \textbf{Ferrocarril Sonora-Baja California (SBC) 2203}, a rare surviving example of an EMD FT diesel-electric locomotive.

Here are the key details:
\begin{itemize}[leftmargin=*, topsep=0pt, itemsep=0pt]
    \item \textbf{Model Introduced:} The EMD FT model line was introduced in \textbf{1939}.
    \item \textbf{Specific Unit History:} This specific locomotive was built in \textbf{1945} and sold to the Mexican railroad SBC in 1964.
\end{itemize}
It is currently preserved at the National Museum of Mexican Railways in Puebla, Mexico.
&
\textbf{50\% Complete} \newline Covers 1 of 2 essential facts. Fails to correctly identify the introduction year.
 &
\textbf{final verdict:} HAS CLEAR CONTRADICTION(S). \newline \textbf{reasoning:} The response claims the model line was introduced in 1939 and the unit built in 1945, which directly contradicts Fact \#2 (introduced in 1949).  \\
\midrule

What genus does this animal belong to? \newline
\includegraphics[width=\linewidth]{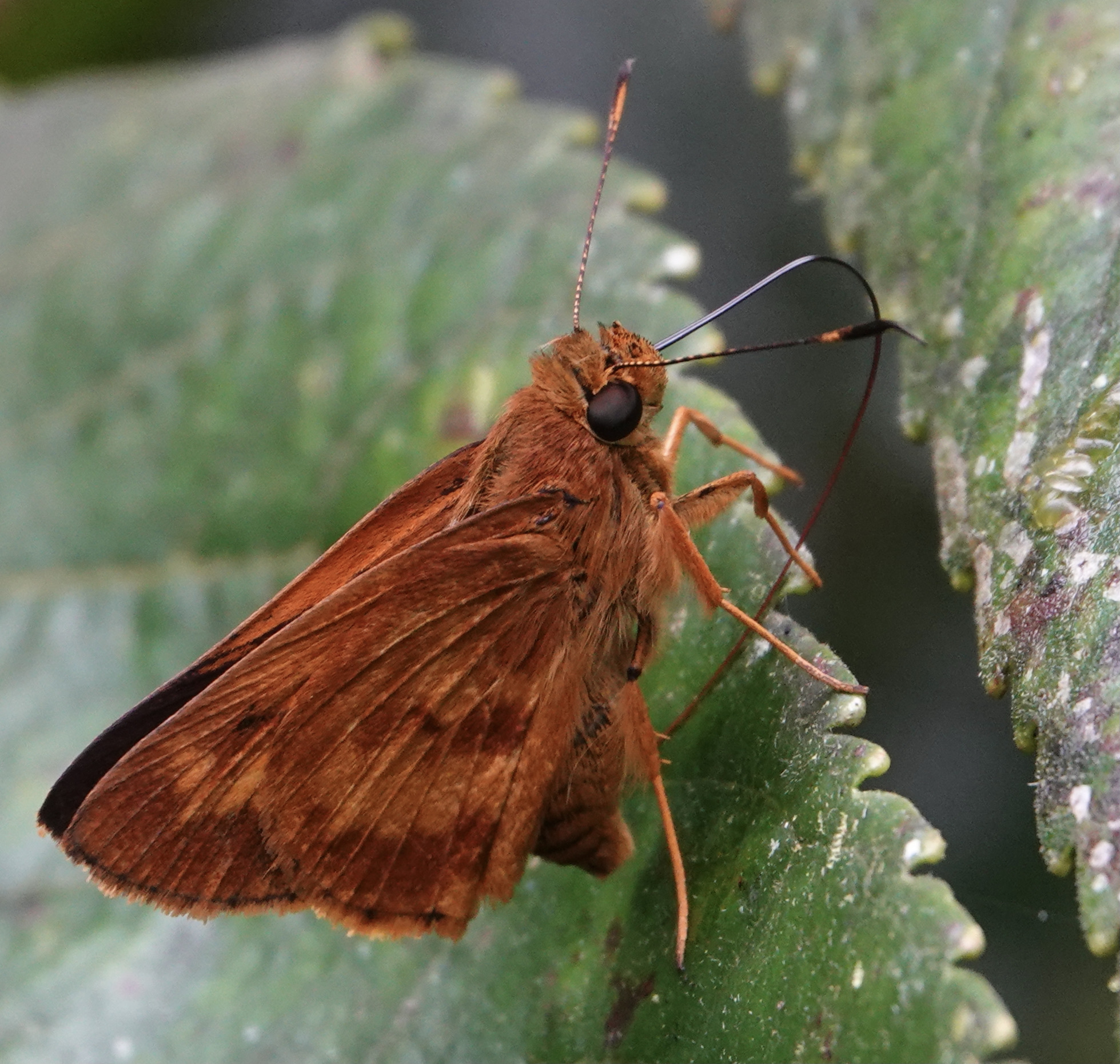}
\tiny{Image: \href{https://www.inaturalist.org/photos/123384584}{Racta apella} by desertnaturalist, \href{https://creativecommons.org/licenses/by/4.0/}{CC BY 4.0}}
&
\begin{enumerate}[label=\arabic*., leftmargin=*, topsep=0pt, partopsep=0pt, itemsep=0pt, parsep=0pt]
    \item[] \textit{\underline{Essential Facts}}
    \item The creature belongs to the genus Racta.
    \item[] \textit{\underline{Non-essential Facts}}
    \item The scientific name for the creature is Racta apella.
    \item The genus Racta is part of the butterfly family Hesperiidae.
    \item The apella skipper is native to South America and Central America.
    \item Skippers are known for their hooked antennae.
\end{enumerate} &
Based on the visual features, this butterfly belongs to the genus \textbf{Thymelicus}.

Here is a breakdown of why:
\begin{itemize}[leftmargin=*, topsep=0pt, itemsep=0pt]
    \item \textbf{Family:} It is clearly a member of the Hesperiidae family, identifiable by the hooked antennae.
    \item \textbf{Genus Characteristics:} The genus Thymelicus is characterized by this specific rusty-orange or golden-brown coloration.
    \item \textbf{Species differentiation:} It is likely either a Small Skipper or an Essex Skipper.
\end{itemize}
Both species are common in \textbf{Europe} and parts of the Middle East.
&
\textbf{0\% Complete} \newline Covers 0 of 1 essential facts. Fails to identify the correct genus.
 &
\textbf{final verdict:} HAS CLEAR CONTRADICTION(S). \newline \textbf{reasoning:} The response identifies the genus as Thymelicus, which contradicts Essential Fact \#1 (Racta). Consequently, it provides incorrect species and location information (Europe vs. South America).  \\
\midrule

Write an elaborative description for the image.
\newline
\includegraphics[width=\linewidth]{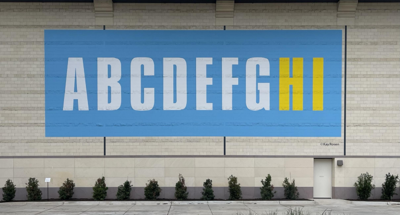}
(The image can be viewed at a larger size \href{https://storage.googleapis.com/docci/data/images_aar/aar_test_04862.jpg}{here.})
&
\begin{enumerate}[label=\arabic*., leftmargin=*, topsep=0pt, partopsep=0pt, itemsep=0pt, parsep=0pt]
    \item[] \textit{\underline{Essential Facts}}
    \item This image contains the first nine letters of the alphabet.
    \item The letters `H' and `I' are painted in yellow, making the word `hi'.
    \item The letters `ABCDEFG' are painted in white.
    \item The painting is on a white brick wall.
    \item[] \textit{\underline{Non-essential Facts}}
    \item At the bottom frame... are shrubs...
    \item The large sidewalk... allows for space to view the mural...
    \item The artist's name, Kay Rosen, appears in black text below the mural...
\end{enumerate} &
This eye-level, outdoor shot captures a large, contemporary mural painted on an exterior brick wall. The mural itself is a vibrant, sky-blue rectangle featuring the capital letters `ABCDEFGHI' in a bold, sans-serif font. The letters `A' through `G' are rendered in stark white, while the letters `H' and `I' are distinctly painted in a vivid yellow, cleverly forming the word `HI' and drawing attention to this friendly greeting. The artist's signature, `© Kay Rosen,' is visible in small white text in the bottom right corner of the blue mural. 
\newline 
(... rest of the response) &
\textbf{100\% Complete} \newline Covers 4 of 4 essential facts. &
\textbf{final verdict:} HAS CLEAR CONTRADICTION(S). \newline \textbf{reasoning:}
The response states, "The artist's signature... is visible in small white text in the bottom right corner of the blue mural." \newline According to Fact \#7, the artist's name appears in "black text below the mural." The response incorrectly describes the color and location of the signature.  \\
\bottomrule
\end{tabular}%
}
\vspace{-12pt}
\end{table*}

The \textit{FACTS Multimodal} benchmark evaluates the ability of models to generate factually accurate text in response to image-based questions. This task requires the integration of visual grounding with parametric world knowledge, a critical capability for modern multimodal systems. The evaluation framework is designed not only to verify the correctness of claims but also to ensure that responses are sufficiently comprehensive to be helpful to the user.
\vspace{-4pt}
\subsection{Data}
\vspace{-4pt}

\begin{figure*}[t!]
    \centering
    \includegraphics[width=1.0\textwidth]
    {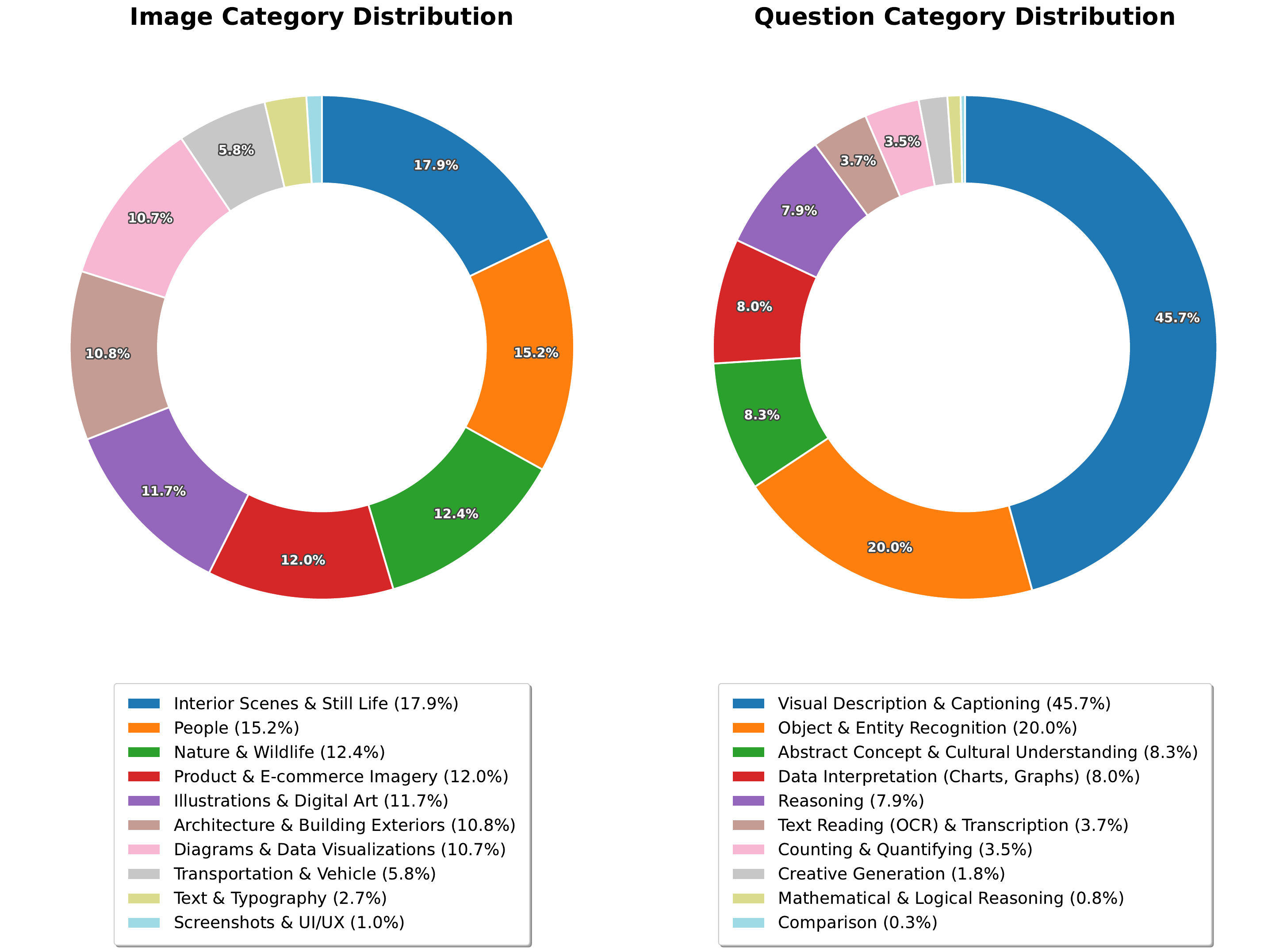}
    
    \caption{Distributions of image and question categories in the FACTS Multimodal benchmark.}
    \label{fig:distr_mm}
\end{figure*}

The evaluation set contains approximately 1,500 questions, divided and filtered into a 711-item \textit{public set} and an 811-item \textit{private set}. Questions were curated from various sources to reflect diverse real-world user queries and were filtered to focus on objective, information-seeking tasks. The benchmark covers a range of capabilities, including detailed visual description, data interpretation from charts and graphs, object recognition, and logical reasoning about visual scenes. Figure \ref{fig:distr_mm} presents the distribution of image and question categories in the public set.

\subsection{Metrics}
The evaluation produces two primary scores --- a \textit{Coverage} score and a \textit{No-Contradiction} score --- determined by an automated judge that evaluates each response against a human-authored, ground-truth rubric. This rubric-based methodology is central to our benchmark, as it provides a scalable and objective framework for evaluation.

\paragraph{Reference Rubric Creation.} For each question, human annotators created a detailed rubric listing relevant facts. Facts that are critical for a complete and satisfactory answer are labeled as \textit{Essential}, while other relevant, contextual facts are labeled as \textit{Non-Essential}. This distinction allows for a nuanced evaluation of both correctness and substance.

\paragraph{Automated Evaluation.} An automated judge, acting as a meticulous fact-checker, is used to verify the model’s response is factual using two boolean verdicts:
\begin{itemize}
\item Coverage verdict: This boolean verdict verifies that the model response includes the essential facts specified in the ground-truth rubric.
\item No-Contradiction verdict: This boolean verdict verifies that the model response does not include any claims that contradict either the ground-truth rubrics (essential and non-essential), common knowledge or the input image itself.
\end{itemize}
Accuracy score: only responses that both cover essential facts and do not include any contradictions are considered accurate. The overall accuracy score is the percentage of such responses in the set. Table~\ref{tab:mm_examples_final} demonstrates this dual-verdict process, showcasing the detailed reasoning the autorater provides to justify its verdicts and the nuances of the errors it can detect.

Table~\ref{tab:mm-results} presents the results of our main FACTS Multimodal benchmark. The Gemini model family is more recall-oriented than other families, demonstrating high Coverage scores. Conversely, GPT models are more precision-oriented, achieving the highest No-Contradiction Scores.

\begin{table}[t]\centering
\caption{Detailed results on the FACTS Multimodal benchmark. Numbers are reported on top of the private and the public set.}\label{tab:mm-results}
\scriptsize
\begin{tabular}{lrrr}\toprule
Model & Accuracy (\%) & No-Contradiction (\%) & Coverage (\%) \\\midrule
Gemini 2.5 Pro & 46.9 & 58.8 & 67.7 \\
Gemini 3 Pro & 46.1 & 57.3 & 68.4 \\
GPT-5 & 44.1 & 64.7 & 59.9 \\
GPT-5.1 & 41.8 & 65.0 & 58.3 \\
GPT-5 mini & 41.5 & 65.1 & 58.8 \\
Gemini 2.5 Flash & 41.0 & 53.9 & 64.8 \\
GPT-4.1 & 40.1 & 62.2 & 57.3 \\
o3 & 39.9 & 55.7 & 62.2 \\
Claude 4.5 Opus & 39.2 & 51.1 & 62.1 \\
o4 mini & 34.5 & 47.6 & 59.1 \\
Claude 4.1 Opus & 33.1 & 49.2 & 54.6 \\
Claude 4.5 Sonnet \textit{Thinking} & 31.1 & 42.6 & 56.0 \\
Claude 4 Sonnet & 28.6 & 46.2 & 50.2 \\
Grok 4 & 25.7 & 32.0 & 57.8 \\
Grok 4 Fast & 17.7 & 22.0 & 50.9 \\
\bottomrule
\end{tabular}
\end{table}

\subsection{Autorater Validation.} The credibility of our automated metrics was established by validating them against human ground-truth annotations.
For \textit{Coverage},  the human annotation task mirrored the autorater's function precisely: given a model response, human raters marked each essential fact from the rubric as either ``supported'' or ``unsupported.'' The autorater was tasked with the same objective. This allowed for a direct comparison of the final ``percentage of essential facts supported'' metric, on which our autorater achieved a high degree of reliability with a Spearman's rank correlation of 0.64 with human judgments. After applying a threshold to convert this to a boolean outcome by using a threshold of 0.5 to ensure most facts are covered, we obtained a macro F1 score of 72.3.  

\begin{table}[t]\centering
\caption{Performance of FACTS Multimodal Autoraters: Coverage and No-Contradiction.}\label{tab:mm_performance_metrics}
%\resizebox{ extwidth}{!}{ % use this if the table is too large
\begin{tabular}{lrr}\toprule
Metric & Coverage & No-Contradiction \\\midrule
Macro F1 & 72.3 & 78.2 \\
Negative F1 & 72.3 & 70.1 \\
Negative Precision & 64.6 & 63.4 \\
Negative Recall & 82.6 & 78.3 \\
Positive F1 & 72.2 & 86.3 \\
Positive Precision & 82.4 & 90.7 \\
Positive Recall & 64.2 & 82.3 \\
\bottomrule
\end{tabular}
\end{table}

\begin{figure*}[t] 
  \centering
  \includegraphics[width=1\textwidth]{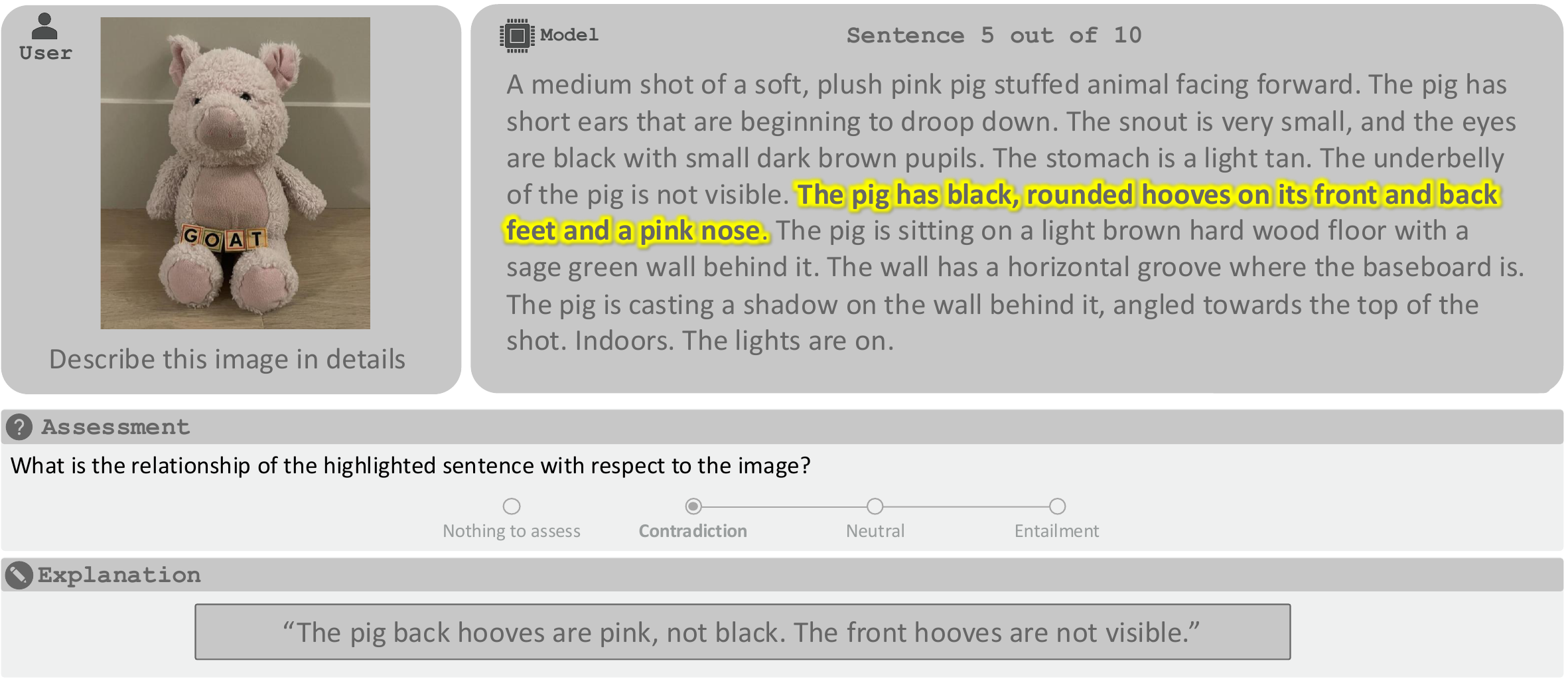}
\caption{Example of the sentence-level annotation interface. Annotators are shown the image, the full model-generated paragraph, and a highlighted sentence. They assess its factuality by selecting a label (here, `Contradiction') and providing a textual explanation for any inaccuracies observed.}
\label{fig:sentence_annotation_template}
\vspace{-15pt}
\end{figure*}

For \textit{No-Contradiction}, we validated the autorater against fine-grained human annotations. The process required a comprehensive review of the entire model response; as illustrated in Figure~\ref{fig:sentence_annotation_template}, annotators assessed the text sentence-by-sentence, using the interface to mark whether each sentence contained a contradiction. This validation achieved a macro F1 score of 78.2. Table~\ref{tab:mm_performance_metrics} details these results, where the positive class denotes the absence of a contradiction.

\section{FACTS Parametric}
\label{sec:searchoff}

The \textit{FACTS Parametric} benchmark assesses the ability of models to accurately answer factual questions (\S\ref{sec:parametric-prop}) that users care about, without the aid of external tools. 
The questions are derived by user-interest, and their answers are confirmed to exist within Wikipedia, a source available for LLM pretraining.
Each of the questions was found challenging through adversarial sampling with a suite of open-source models (see \S\ref{sec:parametric-data-processing}). 
The resulting benchmark measures how well models can recall challenging facts that users care about and that are supported by a primary source. 

\subsection{Data}\label{sec:parametric-data}
FACTS Parametric consists of $2104$  QA pairs, equally divided into a $1052$-item public set and a $1052$-item private set.  
We verified that the underlying intent of each question reflects widespread user interest.
The questions were then subjected to adversarial sampling, and human verification to create a reliable and challenging benchmark (see \S\ref{sec:parametric-data-processing}).

\begin{figure}[t]
    \centering
    % First image
    \begin{minipage}[t]{0.4\textwidth}
        \centering
        \vspace{0pt}
        \includegraphics[width=\linewidth]{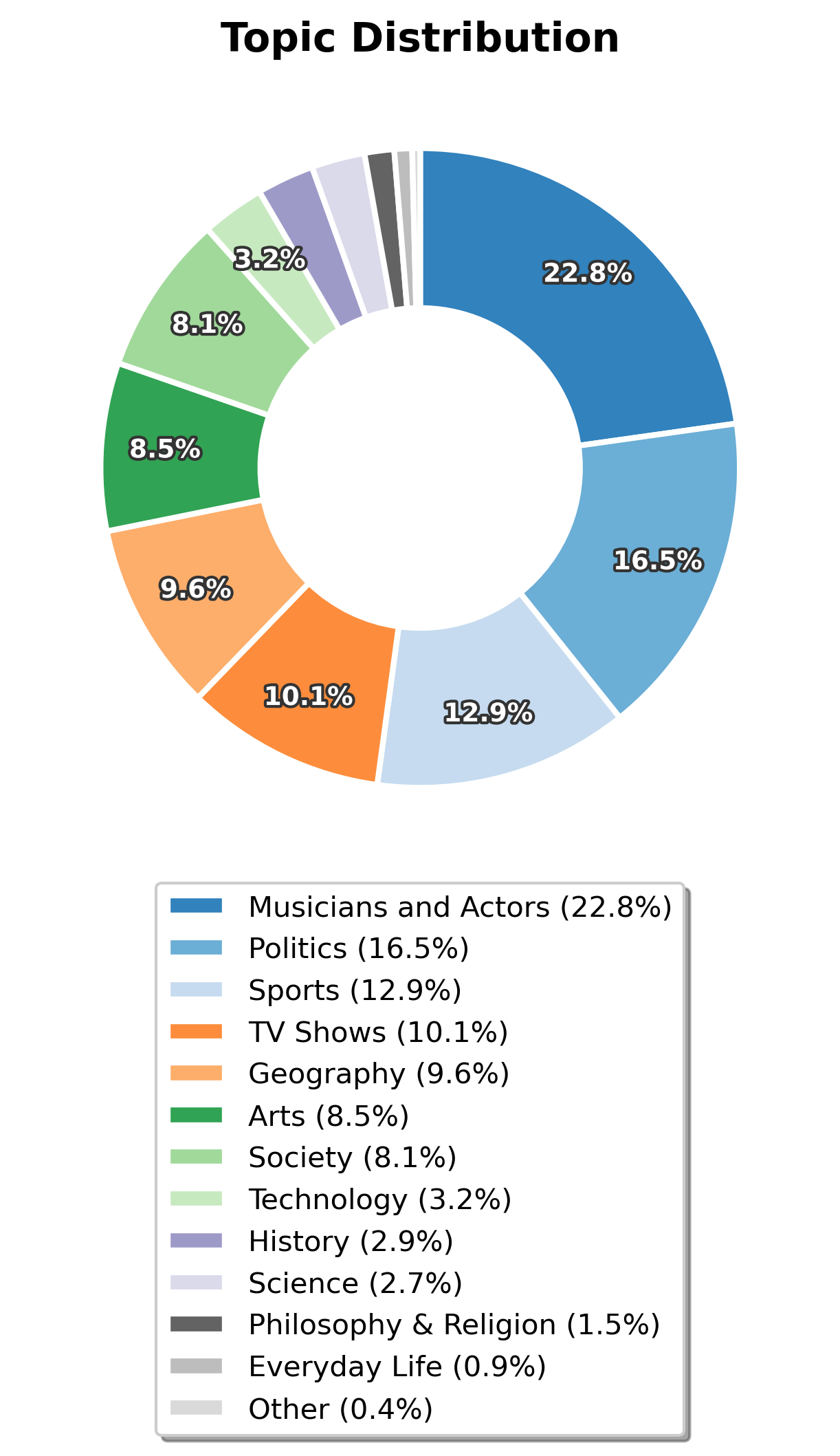}
        \\[0.5ex]
    \end{minipage}
    \begin{minipage}[t]{0.4\textwidth}
        \centering
        \vspace{0pt}
        \includegraphics[width=\linewidth]{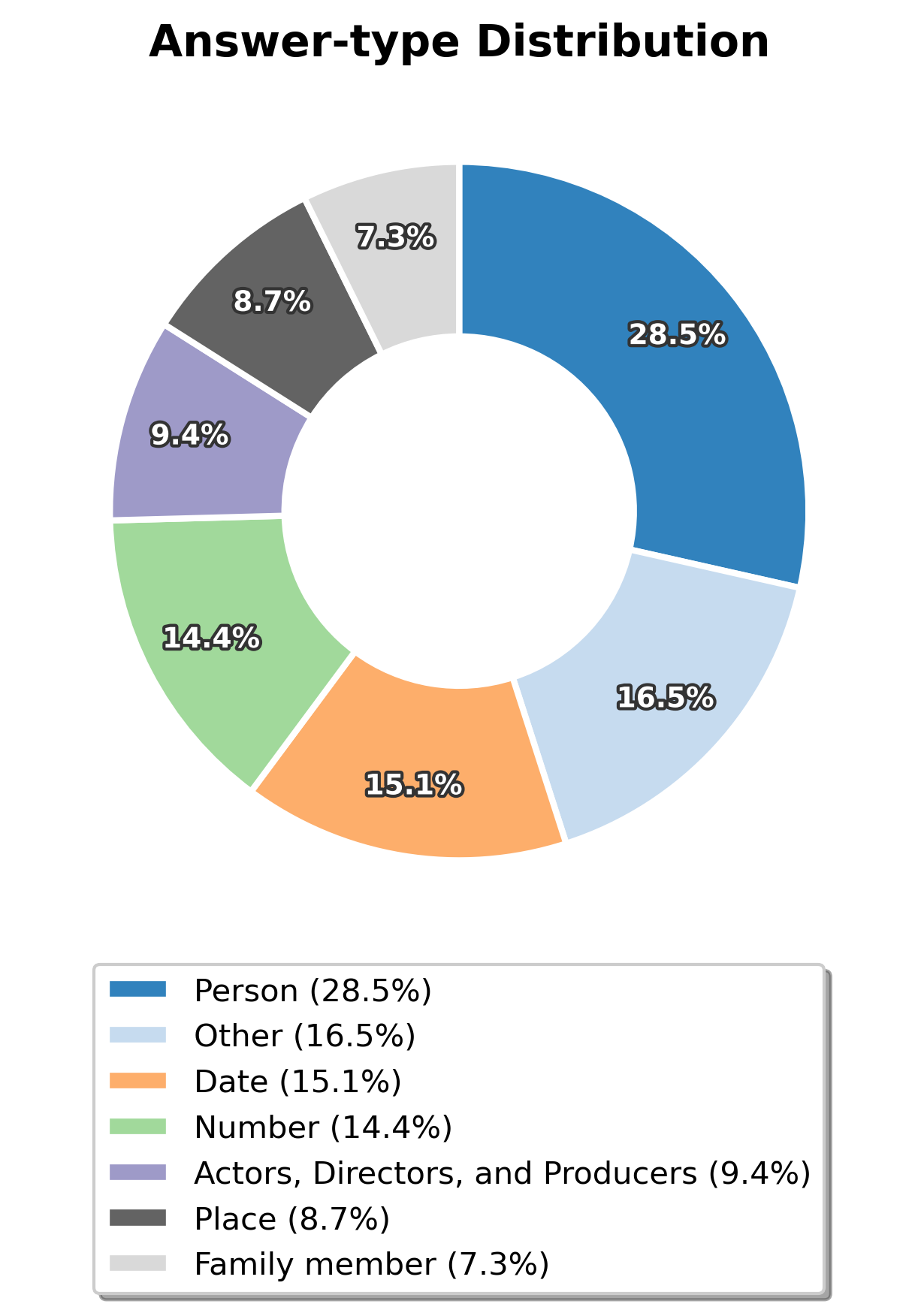}
        \\[0.5ex]
    \end{minipage}
    \caption{Distributions of context domain and of answer type as a percent of the total set of questions in the FACTS Parametric benchmark.}
    \label{fig:distr_parametric}
    \vspace{-15pt}
\end{figure}

The resulting queries span a broad range of topics, including politics, sports, and technology, while the answer types fall into diverse categories such as people, dates, and places. Figure \ref{fig:distr_parametric} presents the breakdown of these topics and answer types in the public set, illustrating the distribution of challenging factoid queries as they appear in real-world traffic. Finally, Table~\ref{tab:examples_parametric} presents a set of examples from the public dataset.

\subsubsection{FACTS Parametric properties}\label{sec:parametric-prop}
\vspace{-5pt}
\paragraph{Questions Reflect User Traffic.} 
One of our goals with FACTS Parametric is to evaluate how familiar models are with facts users genuinely care about. 
To do this, we collected questions that reflect interest shown by many users. However, since strictly following these guidelines tends to yield highly popular (and therefore easier) topics, we deliberately selected the least frequent topics from the eligible set. We then further refined the list of initial questions using adversarial sampling to ensure we only retained questions that remain challenging for models (see \S\ref{sec:parametric-data-processing}).

\vspace{-13pt}
\paragraph{Answers Explicitly Supported by a Prevalent Source.} 
To effectively measure the recall of parametric knowledge acquired during pre-training, we established a key criterion for FACTS Parametric: every answer must be explicitly supported by information found within Wikipedia documents. We selected Wikipedia as the mandatory source because its content is highly prevalent and widely assumed to be a significant component of the training corpora for all LLMs. This constraint helped ensure that the benchmark evaluates knowledge the model was likely exposed to during training, thereby allowing a clearer assessment of its ability to recall factual information learned during that phase, especially when running the assessment with no access to web-search tools.
\vspace{-13pt}
\paragraph{Factoid Criteria.}
To facilitate straightforward and reliable model assessment, we designed FACTS Parametric with a few important properties in mind:
\vspace{-10pt}
\begin{itemize}
    \item \textbf{Single, Atomic Fact:} Each question targets exactly one piece of factual information, avoiding multi-part queries. This ensures that evaluation focuses on recalling one easily verifiable fact.

    \item \textbf{Unambiguous Answer:} Questions are designed to have only one distinct, correct answer, minimizing ambiguity during evaluation.

    \item \textbf{Clear Answer Specification:} The expected type of answer (e.g., person, location, date) or the required granularity is typically stated in the question itself, or is strongly implied.

    \item \textbf{Concise, Factual Answers:} The expected answer is a short entity (like a name, a number, or a specific term) rather than a simple ``yes/no'' response (which the model can 'guess') or a long detailed response. This simplifies matching model outputs to the ground truth.

    \item \textbf{Stable Facts:} The benchmark focuses on facts that are either static (e.g., ``What is the capital of France?'') or explicitly time-anchored within the question (e.g., ``Who was the US president in 1995?''), ensuring the stability and longevity of the ground truth answers.
\end{itemize}

\begin{table*}[t]
\centering
\caption{Example questions from the \textit{FACTS Parametric} benchmark (Public set).}
\label{tab:examples_parametric}
\scriptsize
\ra{1.25}
\setlength{\tabcolsep}{6pt}
\renewcommand{\arraystretch}{1.2}

\begin{tabularx}{0.83\linewidth}{p{4.0cm} p{3.0cm} p{2.0cm} p{1.5cm} p{1.5cm}}
\toprule
\textbf{Question} & \textbf{Answer} & \textbf{URL Suffix} & \textbf{Topic} & \textbf{Answer Type} \\
\midrule
kevin abstract hometown &
Corpus Christi, Texas &
Kevin\_Abstract &
People (artists) &
Place \\ & & & & \\

who played harmonica on the rockford files theme song &
Tommy Morgan &
Tommy\_Morgan &
People (artists) &
Performer \\ & & & & \\

cristian mijares record &
59 wins, 9 losses, and 2 draws &
Cristián\_Mijares &
People (sports) &
Person \\ & & & & \\

triphosgene boiling point (°C) &
206 °C (\emph{acceptable range 203–206}) &
Triphosgene &
Science &
Number \\
\bottomrule
\end{tabularx}
\end{table*}

\subsubsection{Data Processing}\label{sec:parametric-data-processing}

To satisfy the structural requirements detailed in \S\ref{sec:parametric-prop} and construct a challenging benchmark, we implemented a multi-stage filtering pipeline. 
First, we applied automatic LLM-based filters to the initial set of questions, which were collected to reflect user interest, to identify queries satisfying the factoid criteria above.
Next, we utilized an adversarial sampling mechanism to isolate the most challenging examples. Finally, we conducted human verification to confirm adherence to all specified properties.

\vspace{-10pt}
\paragraph{Adversarial Sampling with Open LLMs.}
While FACTS Parametric is grounded in questions frequently asked by users, a key goal is to ensure the benchmark remains challenging for frontier LLMs, avoiding saturation in the near future. To achieve this, we employ an adversarial sampling strategy during the data collection phase.

First, we generate preliminary ``silver'' labels for all questions using \textit{Gemini-2.5-Pro} equipped with search tools. In this setup, the model returns both a generated answer and the specific search results used to derive it. Since we aim to ground our benchmark in verifiable sources, we filter this set to retain only the questions where the model's answer is supported by a Wikipedia URL found in the search results.

Next, to identify the most challenging questions among this verified set, we collect responses from five strong open-weight models (i.e., models with publicly available weights that can be run locally).
We specifically utilized open-weight architectures to decouple the adversarial selection process from the proprietary API models used in our evaluation, thereby ensuring unbiased filtering. Crucially, we query these models in a closed-book setting, without access to external search tools. 

Finally, we retain only the questions that none of these open-weight models answered correctly.\footnote{At this phase, ``correctly'' is defined as matching the silver answer provided by Gemini-2.5-Pro. Note that these are preliminary filters; the final labels are determined later.} As a final step, this refined list is sent to human annotation for rigorous verification (see below).

\paragraph{Human Annotation.}
To maintain data quality and validity, each question-answer (QA) pair was verified by three independent third-party human annotators. Annotators received the question, the LLM-generated candidate answer (silver label), and a supporting URL from the initial data collection. Their role was to confirm the factual accuracy of the QA pair and its compliance with the FACTS Parametric properties detailed in Section~\ref{sec:parametric-prop}.

Specifically, annotators performed the following evaluations for each example:
\vspace{-10pt}
\begin{itemize}
    \item \textbf{Factuality Assessment:} Determine the correctness of the provided answer for the given question. Annotators were explicitly instructed to base their judgment on information available on the web, and not solely on the provided URL. The possible verdicts were: 
    \begin{itemize}
        \item \textit{Accurate:} The answer is verifiably correct based on accessible online information.

        \item \textit{Inaccurate:} The answer is deemed incorrect due to one of the following reasons: (a) reliable online sources provide contradictory information, (b) no supporting information could be found online, or (c) the topic is disputed, with both supporting and contradicting information found online.
        
    \end{itemize}
    
    \item \textbf{Properties Compliance Check:} Verify whether the question and answer strictly adhere to all the dataset properties described in Section~\ref{sec:parametric-prop}. The outcome was a simple True/False verdict.
    
    \item \textbf{Wikipedia Evidence Extraction:} Identify and provide a Wikipedia document that confirms the fact represented by the QA pair. The rater was given a Wikipedia URL that might contain the required information, but it had to be checked. When annotators did not find a supporting Wikipedia URL, the example was disqualified from appearing in the final dataset.

    \item \textbf{Correction Provision:} If a QA pair was initially rated as `Inaccurate' or failed the `Properties Compliance Check', but the issue could be resolved with a simple correction (to either the question, answer, or URL) without altering the original user intent, annotators were instructed to provide the corrected version. Otherwise, the QA pair was discarded. In the cases where a correction was proposed, a new annotator was instructed to finalize the examples with proposed correction. The extra annotator had the option to disqualify the examples from appearing in the final set.
    
\end{itemize}

This multifaceted annotation process, including verification, evidence extraction, and correction, was vital for constructing a high-quality benchmark.

\begin{table}[t]\centering
\caption{Detailed results on the FACTS Parametric benchmark. Numbers are reported on top of the private and the public set.}\label{tab:parametric-results}
%\resizebox{ extwidth}{!}{ % use this if the table is too large
\scriptsize
\begin{tabular}{lrrrr}\toprule
Model             &F1     &Accuracy (\%) &Attempted accuracy (\%)& Hedging rate (\%) \\\midrule
Gemini 3 Pro      &$77.0$ &$76.4$ &$77.6$ & $1.4$ \\
Gemini 2.5 Pro    &$63.8$ &$63.2$ &$64.5$ & $1.9$ \\
Grok 4            &$60.9$ &$58.5$ &$63.6$ & $7.9$ \\
GPT-5             &$59.7$ &$55.7$ &$64.3$ & $13.3$ \\
o3            &$57.6$ &$57.0$ &$58.2$ & $1.9$ \\
GPT-4.1           &$52.5$ &$51.5$ &$53.6$ & $3.8$ \\
GPT-5.1           &$45.8$ &$43.2$ &$48.7$ & $11.3$ \\
Claude 4.1 Opus  &$39.0$ &$33.2$ &$47.4$ & $29.9$ \\
Claude 4.5 Sonnet &$34.3$ &$28.9$ &$42.2$ & $31.3$ \\
Claude 4.5 Opus  &$32.8$ &$30.5$ &$35.5$ & $14.0$ \\
Gemini 2.5 Flash  &$31.4$ &$30.6$ &$32.3$ & $5.1$ \\
Claude 4 Sonnet   &$26.3$ &$20.3$ &$37.2$ & $45.1$ \\
GPT-5 mini        &$24.2$ &$16.0$ &$49.6$ & $67.6$ \\
o4 mini       &$21.6$ &$20.4$ &$22.9$ & $10.5$ \\
Grok 4 Fast       &$16.9$ &$15.7$ &$18.4$ & $14.4$ \\
\bottomrule
\end{tabular}
\end{table}

\subsection{Metrics}\label{sec:parametric-grader}

We follow the grading scheme proposed by \cite{wei2024measuring} with some modifications. First, we slightly change the examples given in the grader instruction prompt to better represent the scenarios we see in our data. Then, we introduce an additional grading label, \textit{unknown}, which represents cases where the grader is unsure whether the gold answer and the model response are aligned. We find this label to improve the already high accuracy presented by the grader. 

Accordingly, the resulting grader automatically grades each model response as either \textit{correct, incorrect, not-attempted, or unknown}. Our primary metric is \textit{accuracy}, which is measured by the percentage of correct responses. We also report three secondary metrics: \textit{Hedging rate} (the percentage of not-attempted), \textit{attempted-accuracy}, and \textit{F1-score}, the harmonic mean of accuracy and accuracy given attempted.

To enhance grading reliability, we sample three grades from \textit{Gemini-2.5-Pro} for each {query, gold-answer, response} triplet and average them to determine the final score.
We observed that utilizing a powerful model for grading improves grading accuracy. We ultimately decided to standardize on \textit{Gemini-2.5-Pro} as our sole judge to keep the benchmark simple and easy to maintain.
We validated this choice by comparing it against a mixed-model panel (sampling once each from \textit{Gemini-2.5-Pro}, \textit{GPT-o3}, and \textit{Grok-4}). The results confirmed that using \textit{Gemini-2.5-Pro} alone preserves the same relative performance trends and rankings as the more complex ensemble.

Table \ref{tab:parametric-results} presents the main results for the FACTS-parametric benchmark. For all metrics, we report the average score derived from three sampled grades per model response. These results illuminate model behavior regarding uncertainty: while guessing is the optimal strategy for standard accuracy, the ``attempted-accuracy'' metric incentivizes hedging.

This trade-off is evident when comparing GPT-o3 and GPT-5. Although GPT-o3 achieves higher raw accuracy ($57.0\%$ vs. $55.7\%$), GPT-5 hedges significantly more often ($13.3\%$ of cases vs. $1.9\%$). Consequently, GPT-5 achieves superior attempted accuracy ($64.3\%$ vs. $58.2\%$) and F1 scores ($59.7$ vs. $57.6$).

\newcommand{\trapname}{Hard Tail}
\newcommand{\harderfactoidname}{Wiki Two-Hop}
\newcommand{\coralname}{Wiki Multi-Doc}
\newcommand{\kraftname}{KG Hops}

\begin{table*}[t]
\centering
\caption{Examples from \textit{FACTS Search} (Public set).}
\label{tab:examples_search_on}
\scriptsize
\ra{1.25}
\setlength{\tabcolsep}{6pt}
\renewcommand{\arraystretch}{1.2}
\begin{tabularx}{0.8\textwidth} { 
   >{\raggedright\arraybackslash}X 
   >{\raggedright\arraybackslash}p{8cm} 
   >{\raggedleft\arraybackslash}X  } \toprule 
\textbf{Slice} & \textbf{Question} & \textbf{Answer}   \\
\midrule
\trapname{} & For the person who had the most followed Instagram account in 2017, how many solo studio albums did they release prior to this accomplishment? &
Two  \\ & \\
\harderfactoidname{} & In which month of 2017 did the European Medicines Agency grant orphan drug designation to the drug with DrugBank ID DB11978? & October \\ & \\
\coralname{} &  What is the sum of the birth years of the British boxer who defeated Vazik Kazarian at the 1960 Summer Olympics, the Moroccan boxer who also competed in the men's light welterweight event at those same Olympics, and the Danish boxer who competed in both the 1960 and 1964 Summer Olympics?  & 5821 \\ & \\
\kraftname{} & Among all the films written by the creator of the TV Program The Sopranos, which one was released the earliest? & Grave of the Vampire \\ & \\
\bottomrule
\end{tabularx}
\end{table*}

\section{FACTS Search}
\label{sec:searchon}
The \textit{FACTS Search} benchmark evaluates the ability of models to use web search. Indeed, most recent generative models are designed to make use of a search tool, and several benchmarks have been proposed to test this capability \cite[e.g., see][]{wei2025browsecomp,yang2024crag}. 

The key challenge when designing such benchmarks is to collect questions whose solution requires search tools. Generally, the only case where this is guaranteed is when the information sought by the question does not appear in the model training data, and does appear in the search results. However, this is hard to guarantee across models (whose training data and cutoff dates are different), and thus we do not pursue this direction. Instead, we focus on other aspects that are hard for models that do not have access to web search. These include tail-entities which are often not sufficiently encoded in parameters, and multi-hop queries (where it's less likely that all hops are encoded in the parameters).

\subsection{Data}

The evaluation set contains 1884 questions, which are split into public and private test sets, of sizes 890 and 994 respectively. The questions were collected from various sources, with the goal of capturing diverse aspects of queries that are likely to require search. Table \ref{tab:examples_search_on} shows some examples from the dataset. 
The set of questions consists of four subsets, each generated using a different strategy.
One subset was written by human raters, and the other three were synthetically generated. See details below.

\begin{table}[t]\centering
\caption{Detailed results on the FACTS Search benchmark. Numbers are reported on top of the private and the public set.}\label{tab:search-results}
\vspace{-5pt}
\scriptsize
\begin{tabular}{lrrrrr}\toprule
Model             &F1     &Accuracy (\%) &Attempted accuracy (\%)& Hedging rate (\%) & Average searches \\\midrule
Gemini 3 Pro      & $85.6$ & $83.8$ & $87.6$ & $4.4$  & $3.39$ \\
GPT-5             & $81.4$ & $77.7$ & $85.5$ & $9.1$  & $4.28$ \\
Claude 4.5 Opus  & $80.0$ & $73.2$ & $88.2$ & $16.9$ & $3.98$ \\
Claude 4.5 Sonnet & $78.5$ & $69.8$ & $89.7$ & $22.2$ & $4.02$ \\
o3            & $77.6$ & $74.8$ & $80.7$ & $7.3$  & $4.64$ \\
Grok 4            & $77.4$ & $75.3$ & $79.6$ & $5.4$  & $4.5$ \\
GPT-5 mini        & $76.5$ & $67.9$ & $87.5$ & $22.5$ & $3.97$ \\
Grok 4 Fast       & $75.1$ & $67.3$ & $85.0$ & $20.9$ & $4.74$ \\
Claude 4.1 Opus  & $75.1$ & $65.0$ & $88.9$ & $26.9$ & $4.66$ \\
Claude 4 Sonnet   & $74.7$ & $66.3$ & $85.7$ & $22.6$ & $4.49$ \\
GPT-4.1           & $65.8$ & $64.6$ & $67.1$ & $3.7$  & $3.17$ \\
o4 mini       & $68.7$ & $66.2$ & $71.4$ & $7.2$  & $4.66$ \\
Gemini 2.5 Pro    & $68.2$ & $63.9$ & $73.1$ & $12.6$ & $3.64$ \\
Gemini 2.5 Flash  & $67.2$ & $60.0$ & $76.5$ & $21.5$ & $3.38$ \\
GPT-5.1           & $66.7$ & $62.4$ & $71.7$ & $13.0$ & $3.34$ \\
\bottomrule
\end{tabular}
\vspace{-5pt}
\end{table}

 \begin{itemize}
     \item {\bf \trapname{}} : 
     % A set of 328 questions written by human raters. 
     A set of questions written by human raters as follows. 
     The raters were instructed to write questions that require information that is challenging to extract with web search. Namely, that there is no single-step web search answer available on the first page, or the information is not readily available as a verbatim piece of text on the internet. Raters were also asked to verify that the Gemini model publicly available at the time (Gemini 1.5) could not solve these, even when using search.
     \item {\bf \harderfactoidname{}} - 
     %A set of 932 questions, generated synthetically as follows. 
     A set of question-answer pairs, synthetically generated using Wikipedia as follows. 
     An initial set of QA pairs was extracted from Wikipedia abstracts, and filtered to focus on tail entities. Next, each question was modified to be a harder, multi-step question via synthetic alteration. This was achieved by substituting the main entity of the question with a different description of this entity that is extracted from the Google Knowledge Graph. For example, ``What is the birthplace of John Lennon'' could be modified to ``What is the birthplace of Yoko Ono’s spouse''. These questions were not adversarially filtered, but evaluation on Search-Off and Search-On Gemini models available at the time gave low accuracies of 30\% and 38\%.
    \item {\bf \coralname{}}  -  
    %A set of 268 questions generated from Wikipedia as follows. 
    A set of question-answer pairs synthetically generated from multiple Wikipedia documents as follows. 
    First, a seed document $D_{seed}$ was sampled. Then, it was used to sample a set of similar documents $D_{related}$, and only the documents $D_{torso}$ with rank that is neither too low nor too high are kept.\footnote{This eliminates documents that are too similar or too distant from $D_{seed}$.} Finally, Gemini was prompted to synthesize a query-answer pair $(Q, A)$ from the content of these $n$ documents. The query $Q$ was formulated to be answerable only by synthesizing information present in both the seed document $D_{seed}$ and one or more documents from the $D_{torso}$ subset. The prompt also encouraged Gemini to find interesting ways to connect information rather than relying on a simple direct chain or a combination of unrelated queries. Next, questions were filtered as follows.
    First, an automated critic model filtered out pairs where the question was not self-contained or the answer was not strictly grounded in the source documents. Second, a hardness filter was applied, discarding any queries that Gemini could correctly answer when utilizing standard web search tools.
    \item {\bf \kraftname{}}  - 
    A set of question-answer pairs synthetically generated using multiple hops in the Google Knowledge Graph.
    To generate these, we first collected common path-queries, such as ``films that actor X appeared in''. These queries were then concatenated and combined with other functions (e.g., max) to create more complex ones. For example, ``films that actor X appeared in'' and ``publication date of film X'' could be combined to 
    create ``publication date of the first film that actor X appeared in''. 
 \end{itemize}
%}

%\subsubsection{Ground Truth Labels}
The collection process above resulted in a set of questions with corresponding gold answers, either written by human raters or extracted automatically as part of the data generated process. In order to further check the quality of the answer for all questions, three independent human raters were asked to rate each question and answer according to the following criteria:
\begin{itemize}
    \item {\bf Correctness}: Use Google search to check that the provided answer is a correct response to the question. 
    \item {\bf Uniqueness}: Check whether there are entities different and distinct from the provided answer that could also be correct answers to the given question.  
    \item {\bf Immutability}: Identify whether the answer to the given query is likely to change in the next five years. 
\end{itemize}
The dataset was filtered to include only questions that all three raters marked as correct, unique and immutable. Finally, all questions were filtered to exclude those that Gemini 2.5 Flash without search answered correctly, thus emphasizing the need for a search tool. The resulting final dataset sizes were 328, 932, 268, 356 for \trapname{},\harderfactoidname{},\coralname{}, and \kraftname{} respectively. 

\subsection{Search Engine}
The goal of the FACTS Search benchmark is to evaluate how well LLMs use search tools. Because performance relies heavily on the specific tool used, meaningful comparison requires that all models access the same search engine. The FACTS leaderboard evaluation uses the Brave Search API as the search tool. All evaluated models receive the same description of the tool. When an LLM triggers a tool call, the API is queried, and the output is appended to the LLM context.
%\amirg{add prompt}

\subsection{Metrics}
To evaluate the quality of model responses, we use a prompted auto-rater. Specifically, given a query, a model response, and a gold response, Gemini 2.0 Flash is prompted to assess if the response is correct, incorrect or does not attempt to answer the query.

Table~\ref{tab:search-results} presents our main FACTS Search results. Notably, Gemini 3 Pro, the highest-performing model, conducts fewer searches on average than other top models, whereas the Grok model family searches the most. Consistent with observations in the FACTS Parametric slice, Claude appears to prioritize accuracy on attempted queries over overall accuracy. 

\begin{table*}[t]\centering
\caption{Examples from \textit{FACTS Grounding v1} (Public set).}
\label{tab:examples}
\scriptsize
% \rowcolors{1}{}{lightgray} 
\ra{1.3}
\resizebox{0.99\linewidth}{!}{
\begin{tabular}{
p{5.1cm}
p{4.5cm}
c
p{5.0cm}
}
\toprule
\textbf{System Instruction} & \textbf{Context Document Description} & \textbf{Context Tokens} & \textbf{User Request} \\ \midrule
 Answer the question using only the information provided in the context. Do not rely on external knowledge or sources. &  The development and deployment of an autonomous robotic system designed to clean skyscraper windows, highlighting its technological advancements, safety implications, and potential impact on the window-washing industry. & $\sim$1.1k &  My sister and her dog live in NYC. I've visited there and have always been fascinated with their tall buildings. Then I thought...someone has to clean those! Then next thing you know, window washing robotos popped up on my feed. How do these robots work? Also what does this mean for the people who do those jobs? \\ \\
 Provide a response based solely on the information provided in the prompt. External sources and prior knowledge must not be used. &  Legal interpretations and effects of the medical marijuana appropriations rider on federal marijuana prosecutions, focusing on differing circuit court approaches to determining compliance with state medical marijuana laws. & $\sim$1.6k &  What did the first circuit conclude?  \\ \\
 This task requires you to answer questions based solely on the information provided in the prompt. You are not allowed to use any external resources or prior knowledge. Present your answer in headed sections with an explanation for each section. Each explanation should be in bullet points with exactly three bullet points. &  Comparison of different economic systems, including free market, command, and mixed economies, highlighting their key characteristics, advantages, and disadvantages. & $\sim$0.9k &  which famous economists are mentioned? \\ \\
 Answer the question based solely on the information provided in the passage. Do not use any external knowledge or resources. &  A study that investigates the correlation between advanced maternal age (40+) and increased risk of obstetric, fetal, and neonatal complications compared to women aged 25-35. & $\sim$2.1k &  Researchers at Foch Hospital in France published this study of pregnancy outcomes in two groups of patients. Please summarize outcomes across the three kinds of complications that the researchers studied.
\\ \bottomrule
\end{tabular}
}
\end{table*}

\section{FACTS Grounding v2}
\label{sec:grounding}
The \textit{FACTS Grounding v2} benchmark extends the \textit{FACTS Grounding} benchmark (referred to as FACTS Grounding v1 below)  previously introduced in \citet{facts_grounding}. \textit{FACTS Grounding} v1 evaluated whether a model response is consistent with the given context document and user query about that document. Here we briefly re-introduce \textit{FACTS Grounding v1}, and then describe the new version, in which the judge models were updated.
Please refer to \citet{facts_grounding} for more details about the original \textit{FACTS Grounding} leaderboard.

\begin{figure}[t]
    \centering
    \begin{minipage}[t]{0.38 \textwidth}
        \centering
        \vspace{0pt}
        \includegraphics[width=\linewidth]{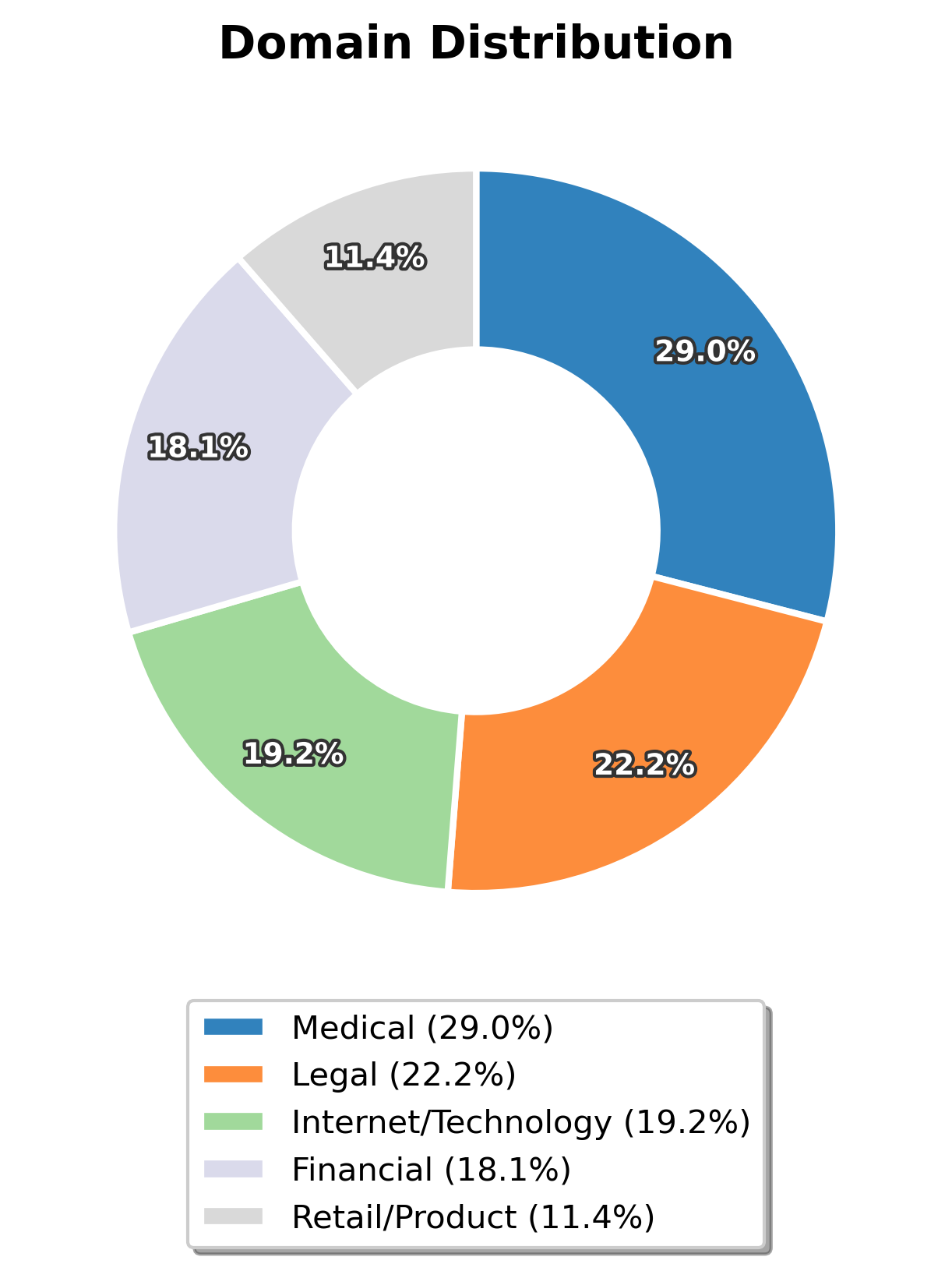}
        \\[0.5ex]
    \end{minipage}
    \begin{minipage}[t]{0.38\textwidth}
        \centering
        \vspace{0pt}
        \includegraphics[width=\linewidth]{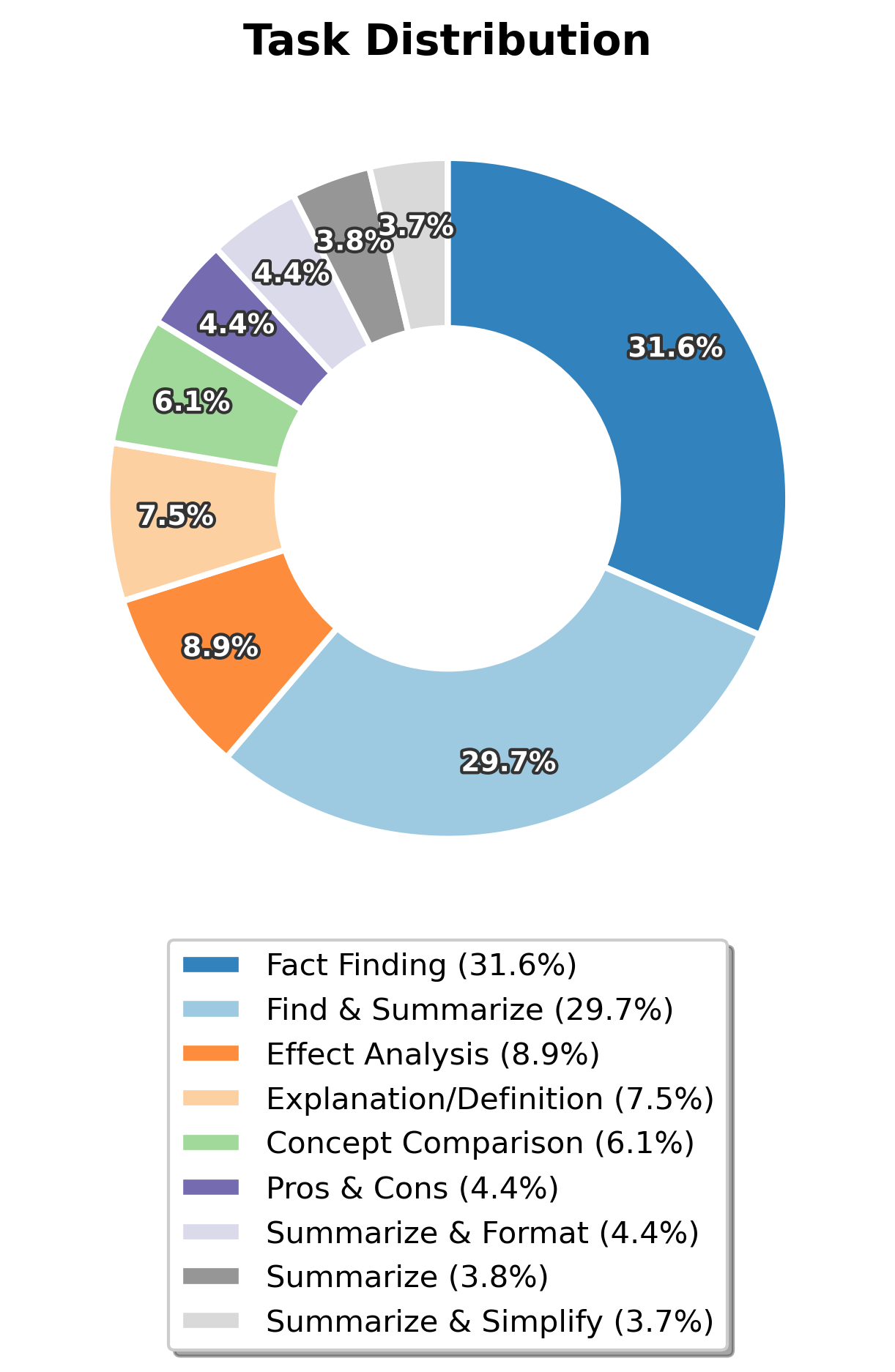}
        \\[0.5ex]
    \end{minipage}
    \vspace{-5pt}
    \caption{Distributions of context domain and of task requested by the user as a percent of the total set of prompts in the benchmark.}
    \label{fig:distr}
    \vspace{-10pt}
\end{figure}

\subsection{Data}

The set of prompts for FACTS Grounding v2 is the same as v1. We provide a description below for completeness.
Third-party human raters were instructed to design prompts requiring the processing of long-form input and the writing of long-form output. These tasks include Q\&A, summarization, and document rewriting. Each example within our evaluation set consists of a context, which is a document or set of reviews sourced from the web, paired with a non-trivial user request that can be addressed using the provided context, necessitating a long-form response. Additionally, each example includes a system instruction directing the model to generate its response exclusively from the given context, without incorporating external knowledge.

To ensure the diversity of the evaluation set, prompts were generated across a range of document lengths (up to 32k tokens) and various enterprise domains, including finance, technology, retail, medical, and legal. The annotation instructions were carefully designed to avoid prompts requiring creative responses, expert-level domain knowledge, mathematical or logical reasoning, or meta-analysis of the text, such as tone analysis or interpretation of author intent. \Cref{tab:examples} provides concrete examples of data instances in the collection. The specific distributions of enterprise domains and of tasks requested by users are shown in Figure \ref{fig:distr}.

\subsection{Metrics}
We keep the same evaluation approach as in FACTS Grounding v1 \citep{facts_grounding}. Specifically, we first prompt multiple ``judge'' LLMs to determine if the response is grounded in the input. Finally, we find ``ineligible'' responses that do not sufficiently address the user request, and mark these as inaccurate, so only grounded and eligible responses are labeled as accurate.
The main updates in FACTS Grounding v2 are the LLMs used as judges, and the prompt used. See details below.

\vspace{-10pt}
\paragraph{Unadjusted Factuality Score.}
As in FACTS Grounding v1, the principal component of our evaluation process is an unadjusted \textit{factuality score}, which is the initial score \textit{before} the adjustment for ineligible responses (described later).

First, we utilize a language model judge to produce a binary classification label identifying whether a full model response is grounded in the user request and the context document given an instruction (see Table~\ref{tab:examples}). A model response is marked with a positive label (``accurate'') if all the claims in the response are grounded in the contents of the prompt, or do not require grounding;  the response is marked with a negative label (``not accurate'') if at least one claim that bears information is deemed to be not grounded in the contents of the prompt. We use two different judge models in order to reduce the bias of a particular judge model, as models have been shown to be biased towards favorably judging their own outputs~\citep{wataoka2024selfpreferencebiasllmasajudge}. The judge models are \textit{Gemini 2.5 Flash}~\citep{comanici2025gemini25pushingfrontier} and \textit{GPT-5}~\citep{openai2025gpt5card}.
We note that \textit{FACTS Grounding v1} used a different set of models: \textit{Gemini 1.5 Pro}~\citep{team2023gemini}, \textit{GPT-4o}~\citep{openai2024gpt4}, and \textit{Claude 3.5 Sonnet}~\citep{anthropic_claude_2024}.

To evaluate the quality of the new judge models, we compared them to human-ratings on a held-out evaluation set (N=320). We also investigated changes to the judge prompt templates. Specifically, we considered two prompt variants: the one from FACTS Grounding v1, and a slightly modified version v2. 
Results are shown in~\cref{tab:grounding-judge-eval}, and demonstrate that the new models combined with the v2 prompt outperform other model-prompt combinations. Given the two judges, the individual factuality score for each judge is the percentage of accurate responses, and the unadjusted factuality score is the average of all judge scores.

\begin{table}[tp]\centering
\scriptsize
\caption{Evaluation of different judge models and evaluation prompts on a private test-set (N=320, class ratio 79:19). Chosen judges via Macro-F$_1$ in \textbf{bold}. }\label{tab:grounding-judge-eval}
\begin{tabular}{llrrrrrrr}\toprule
\textbf{Judge Model} &\textbf{Prompt} & \textbf{Macro-F$_1$} &\textbf{F$_1$ ($+$)}  &\textbf{F$_1$ ($-$)}  &\textbf{Acc.} &\textbf{Prec.} &\textbf{Rec.} \\\midrule
\textbf{gemini-2.5-flash} &\textbf{v2} &\textbf{65.33} &84.51 &46.15 &75.94 &85.71 &83.33 \\
\textbf{gpt-5-2025-08-07} &\textbf{v2} &\textbf{65.18} &80.09 &50.27 &71.56 &89.27 &72.62 \\
gemini-2.5-pro &v1 &64.87 &87.92 &41.82 &80.00 &83.81 &92.46 \\
gemini-2.5-pro &v2 &64.07 &84.17 &43.97 &75.31 &85.02 &83.33 \\
gemini-2.5-flash &v1 &63.44 &85.60 &41.27 &76.88 &83.97 &87.3 \\
gemini-2.0-flash &v1 &63.43 &86.86 &40.00 &78.44 &83.52 &90.48 \\
gpt-5-2025-08-07 &v1 &61.83 &87.48 &36.19 &79.06 &82.69 &92.86 \\
gpt-4o &v1 &59.73 &85.28 &34.19 &75.94 &82.29 &88.49 \\
gemini-1.5-pro &v1 &56.89 &89.09 &24.69 &80.94 &81.11 &98.81 \\
gemini-1.5-pro &v2 &55.65 &86.56 &24.74 &77.19 &80.76 &93.25 \\
gemini-2.5-flash &v1 &54.78 &88.13 &21.43 &79.38 &80.59 &97.22 \\
claude-3-5-sonnet-20241022 &v1 &53.15 &79.36 &26.95 &67.81 &80.16 &78.57 \\
gemini-2.0-flash &v2 &51.25 &86.59 &15.91 &76.88 &79.67 &94.84 \\
\bottomrule
\end{tabular}
\vspace{-10pt}
\end{table}

\vspace{-14pt}
\paragraph{Disqualifying Ineligible Responses.}
\label{sec:metrics-filter}
\begin{table*}[t]\centering
\caption{Examples of ineligible responses: these responses, while fully grounded in the context document, fail to address the user request meaningfully and are consequently considered ineligible.}
\label{tab:examples-quality}
\vspace{-5pt}
\scriptsize
% \rowcolors{1}{}{lightgray} 
\ra{1.3}
\resizebox{0.99\linewidth}{!}{
\begin{tabular}
{
p{4cm}
p{2.7cm}
p{2.7cm}
p{4.4cm}
}
\toprule
\textbf{Context Document Description} & \textbf{User Request} & \textbf{Ineligible Response} & \textbf{Rationale} \\ \midrule
 A research report on renewable energy sources, including wind, solar, and hydroelectric power, with specific statistics and case studies. &  Can you summarize the key advantages and disadvantages of wind energy from this document? &  Wind energy is good because it is renewable and clean, but it has some challenges too. &  (1) The response is extremely vague, failing to provide any specific points from the document, such as the cost-effectiveness, geographic limitations, or impacts on wildlife. (2) It does not engage with the query’s focus on key advantages and disadvantages. \\ \\
  A company’s annual financial report, discussing quarterly earnings, expenditures, future investments, and an analysis of the market environment. &  Summarize the main reasons the company’s revenue decreased in Q3. &  The company faced challenges in Q3 that impacted its revenue. &  (1) The response avoids specifying any reasons, such as market trends, increased competition, or operational setbacks, which would likely be in the document. (2) It does not demonstrate an attempt to engage with or extract relevant details. \\ \\
 A historical article on the causes and consequences of the Great Depression. &  What were the main causes of the Great Depression as explained in the document? &  The Great Depression was a difficult time in history with many causes and effects. &   (1) The response provides no substantive information on the causes, such as stock market speculation, bank failures, or trade policies, which were discussed in the document. (2) It ignores the user's explicit focus on the "main causes."
\\ \bottomrule
\end{tabular}
}
\vspace{-10pt}
\end{table*}
Metrics that are focused on evaluating the factuality of the generated text with respect to a context document can be ``hacked'' by ignoring user intent. Namely, by providing shorter responses that evade conveying adequately comprehensive information, even if such content was an important aspect of a user request, it is possible to achieve a high factuality score while not providing a helpful response. See illustrative examples in \Cref{tab:examples-quality}.

We safeguard against such responses by using prompted judge LLMs to determine whether a given generated response sufficiently addresses the user's request. 
The judge LLM is asked to output a binary label indicating the eligibility of the response: either ``eligible,'' signifying that it answers the user request, or ``ineligible,'' otherwise. Ineligible responses are disqualified from factuality evaluation and the final factuality score is adjusted such that ineligible responses are deemed as \textit{inaccurate}. The judge models used are the same two models used for checking grounding.

\vspace{-2pt}

\section{Conclusion}
\label{sec:conclusion}
\vspace{-5pt}
As LLMs improve, existing benchmarks become saturated. It is thus important to introduce benchmarks that challenge current models. Here, we present the FACTS suite, a benchmark where the top performing model has an average accuracy of only $69\%$, leaving considerable headroom for future progress. The FACTS leaderboard results report high-level metrics such as accuracy. However, it will be useful to obtain more fine-grained analysis, studying what affects the hardness of questions. For example, previous work \citep{KandpalDRWR23} has shown that infrequent entities are harder to learn, and it would be interesting to check if this is reflected in FACTS Parametric. Similarly, it would be interesting to study notions of ``tailness'' for FACTS Search, where some facts might be harder to search for than others. 

Each of the FACTS subsets requires different capabilities to solve, and can serve to drive research on these fronts. FACTS Multimodal requires integration of image understanding with knowledge not represented in the image; FACTS Parametric relies on representing broad factual knowledge in the model parameters, FACTS Search involves effective use of search tools, and FACTS Grounding focuses on the ability to ground the response to context. These are all facets of factuality, where the model relies on different sources of information to generate factual responses.
Naturally, there are aspects of factuality not covered by FACTS, such as video understanding and fast-changing information. In addition, tool-use introduces new factuality challenges, for example when using knowledge-base calls as a tool. We hope FACTS will inspire additional benchmarks that address these areas and others.

\section{Contributions and Acknowledgments}

\begin{itemize}
    \item \textbf{Leads}: Connie Tao, Dipanjan Das, Lukas Haas, Sasha Goldshtein.
    \item \textbf{Benchmark design}: Aileen Cheng, Alon Jacovi, Amir Globerson, Andrew Wang, Chang Liu, Chris Alberti, Eyal Ben-David, Gaurav Singh Tomar, Lukas Haas, Mohamed Amin, Ofir Roval, Prathamesh Bang, Yonatan Bitton, Yulong Yang, Zhongru Wu
    \item \textbf{FACTS Team}: Adam Bloniarz, Aijun Bai, Anfal Siddiqui, Aravindan Raghuveer, Arturo Bajuelos Castillo, Aviel Atias, Ben Golan, Charles Kwong, Corey Fry, Daniel Balle, Deepanway Ghosal, Doron Kukliansky, Dror Marcus, Elena Gribovskaya, Eran Ofek, Honglei Zhuang, Itay Laish, Jan Ackermann, Lily Wang, Meg Risdal, Megan Barnes, Michael Fink, Moran Ambar, Natan Potikha, Nikita Gupta, Nitzan Katz, Noam Velan, Ori Ram, Polina Zablotskaia, Priyanka Agrawal, Rakesh Ghiya, Sanjay Ganapathy, Simon Baumgartner, Sofia Erell, Sushant Prakash, Thibault Sellam, Vikram Rao, Xuanhui Wang, Yaroslav Akulov, Zhen Yang, Zhixin (Lucas) Lai
    \item \textbf{Sponsors}: Koray Kavukcuoglu, Anca Dragan, Avinatan Hassidim, Fernando Pereira, Slav Petrov, Srinivasan Venkatachary, Tulsee Doshi and Yossi Matias who both sponsored the effort and provided technical guidance.
\end{itemize}

We would also like to thank:
\begin{itemize}
    \item \textbf{Gemini Team} for the support and model access.
    \item \textbf{Kaggle Team }for their expertise and releasing the leaderboard.
    \item \textbf{Expert data annotators} who helped to collect examples in the paper.
    \item \textbf{Our reviewers} Lakshman Yagati, John Blitzer, Phoebe Kirk, and Anand Rao for valuable feedback.
\end{itemize}
\label{sec:contributions}

\bibliography{main}

\clearpage

\appendix

% comment

\end{document}